\documentclass[10pt,twocolumn,a4paper]{article}
\usepackage[top=2cm, bottom=2cm, left=2cm, right=2cm]{geometry}

\usepackage{amsmath}
\usepackage{graphicx}
\usepackage{times}
\usepackage{paralist}

\usepackage[sort&compress,numbers]{natbib}

\usepackage{algorithm}
\usepackage[small,bf]{caption}

\usepackage{amsfonts}
\usepackage[keeplastbox]{flushend}
\usepackage{color}
\usepackage[hyphenbreaks]{breakurl}
\definecolor{darkred}{RGB}{153,0,0}
\definecolor{darkblue}{RGB}{0,0,99}
\usepackage[colorlinks=true, linkcolor = darkred,   citecolor = darkred, urlcolor = darkblue]{hyperref}

\usepackage{hyperref}
\usepackage{color, colortbl}
\makeatletter
\def\BState{\State\hskip-\ALG@thistlm}
\makeatother

\definecolor{Gray}{gray}{0.9}

\renewenvironment{thebibliography}[1]{
  \begin{oldthebibliography}{#1}
    \setlength{\itemsep}{0.2em}
    \setlength{\parskip}{0.0em}
}
{
  \end{oldthebibliography}
}

\newcommand{\descr}[1]{\smallskip \noindent \textbf{#1}}

\usepackage{url}
\usepackage[hyphenbreaks]{breakurl}

\makeatletter
\def\url@leostyle{%
  \@ifundefined{selectfont}{\def\UrlFont{}}%
  {\def\UrlFont{}}%
}
\makeatother  
\begin{document}

\title{\bf An Overview of Privacy in Machine Learning}

\author{Emiliano De Cristofaro, UCL \& Alan Turing Institute\\[0.25ex]
e.decristofaro@ucl.ac.uk}

\date{March 2020}

\maketitle

\section{Prologue}\label{sec:intro}

Over the past few years, providers such as Google, Microsoft, and Amazon have started to provide customers with access to software interfaces allowing them to easily embed machine learning tasks into their applications.
Overall, organizations can now use Machine Learning as a Service (MLaaS) engines to outsource complex tasks, e.g., training classifiers, performing predictions, clustering, etc. %
They can also let others query models trained on their data.
Naturally, this approach can also be used (and is often advocated) in other contexts, including government collaborations, citizen science projects, and business-to-business partnerships.

However, if malicious users were able to recover data used to train these models, the resulting information leakage would create serious issues.
Likewise, if the inner parameters of the model are considered proprietary information, then access to the model should not allow an adversary to learn such parameters.
In this document, we set to review privacy challenges in this space, providing a systematic review of the relevant research literature, also exploring possible countermeasures.

More specifically, we provide ample background information on relevant concepts around machine learning and privacy.
Then, we discuss possible adversarial models and settings, cover a wide range of attacks that relate to private and/or sensitive information leakage, and review recent results attempting to defend against such attacks. 

Finally, we conclude with a list of open problems that require more work, including the need for better evaluations, more targeted defenses, and the study of the relation to policy and data protection efforts.

\section{Background}

In this section, we provide some background information and definitions related to machine learning (ML) as well privacy technologies for ML.

\subsection{Machine Learning (ML)}\label{sec:ML}

\descr{Learning process.} ML provides automated methods of analysis for large sets of data, or to perform tasks that are too hard to program by hand.
The general approach to create a machine learning model is as follows~\cite{papernot2018sok}:
\begin{itemize}
\setlength\itemsep{0.1em}
\item {\em Training:} Most ML models can be seen as parametric functions $h_\theta(x)$ taking an input $x$ and a parameter vector $\theta$. 
The input $x$ is often represented as a vector of values called {\em features}.
The space of functions $\{\forall\theta, x \rightarrow h_\theta(x)\}$ is the set of candidate hypotheses to model the distribution from which the dataset was sampled. 
A learning algorithm analyzes the training data to find the value(s) of parameter(s)
$\theta$.
In other words, during training, an ML algorithm aims to learn the ``characteristics'' of some data with respect to a given ``task'' -- e.g., given enough pictures of pets, the algorithm should learn to distinguish what differentiates pictures of, say, cats from dogs.
\item {\em Validation:} The model performance is then validated on a test dataset, which must be disjoint from the training dataset in order to measure the model's generalization. 
In other words, validation performs a sanity check to make sure the algorithm has indeed learned what it is supposed to.
\item {\em Inference (or testing):} Once training completes, the model is deployed to make predictions on inputs unseen during training: the value of parameters $\theta$ are fixed, and the model computes $h_\theta(x)$ for new inputs $x$. 
The model prediction may take different forms but, e.g. for classification, the most common  is a vector assigning a probability for each class of the problem, which
characterizes how likely the input belongs to that class.
In other words, we can now use the algorithm in the wild, and use it for the task at hand -- e.g., given a picture, the algorithm tells us whether that is a picture of a cat or a dog.
\end{itemize}

\descr{Stochastic gradient descent (SGD).} With each function $h$ and $\theta$, one typically associates a loss $L(h,\theta)$, a value that quantifies the cost of any discrepancies between the model's prediction $h_\theta(x)$ and the true value $f(x)$, over all examples $x$ (approximated by the training examples)~\cite{abadi2017protection}. 
Training the model $h$ is the process of searching for a value of $\theta$ with the smallest loss, or with a tolerably small loss. 
Often, both the model $h$ and the loss $L$ are differentiable functions of $\theta$. Therefore, training often relies on a process called stochastic gradient descent (SGD), where one repeatedly picks an example $x$ (or batches thereof), calculates $h_\theta(x)$, and the corresponding loss, and adjusts $\theta$ to reduce the loss by going in the opposite direction of the gradient~\cite{abadi2017protection}. 
In a nutshell, SGD is a very common process used during training to find the settings in the ML algorithm that maximize its accuracy.

\descr{Tasks.} The process discussed above is typical in what is referred to as supervised learning.
However, ML tasks are actually commonly divided into two main types, depending on the structure of the data at hand: 
\begin{itemize}
\setlength\itemsep{0.1em}

\item {\em Supervised learning:} an association between inputs and outputs based on training examples in the form of inputs labeled with corresponding outputs. 
If the output data is categorical, the task is called classification, and real-valued output domains define regression problems~\cite{papernot2018sok}. Classic examples of supervised learning tasks include object recognition in images, machine translation, and spam filtering.
In this setting, the model parameters are adjusted to reduce the gap between model predictions $h_\theta(x)$ and the expected output indicated by the dataset. 
\item {\em Unsupervised learning:} When the method is given unlabeled
inputs, its task is unsupervised. 
Unsupervised learning considers problems such as clustering points according to a similarity metric, dimensionality reduction to project data in lower dimensional subspaces, etc~\cite{papernot2018sok}. 
\end{itemize}
Other types of techniques include {\em reinforcement learning}, which we omit to ease presentation.

\begin{figure}[t]
\centering
\includegraphics[width=0.99\columnwidth]{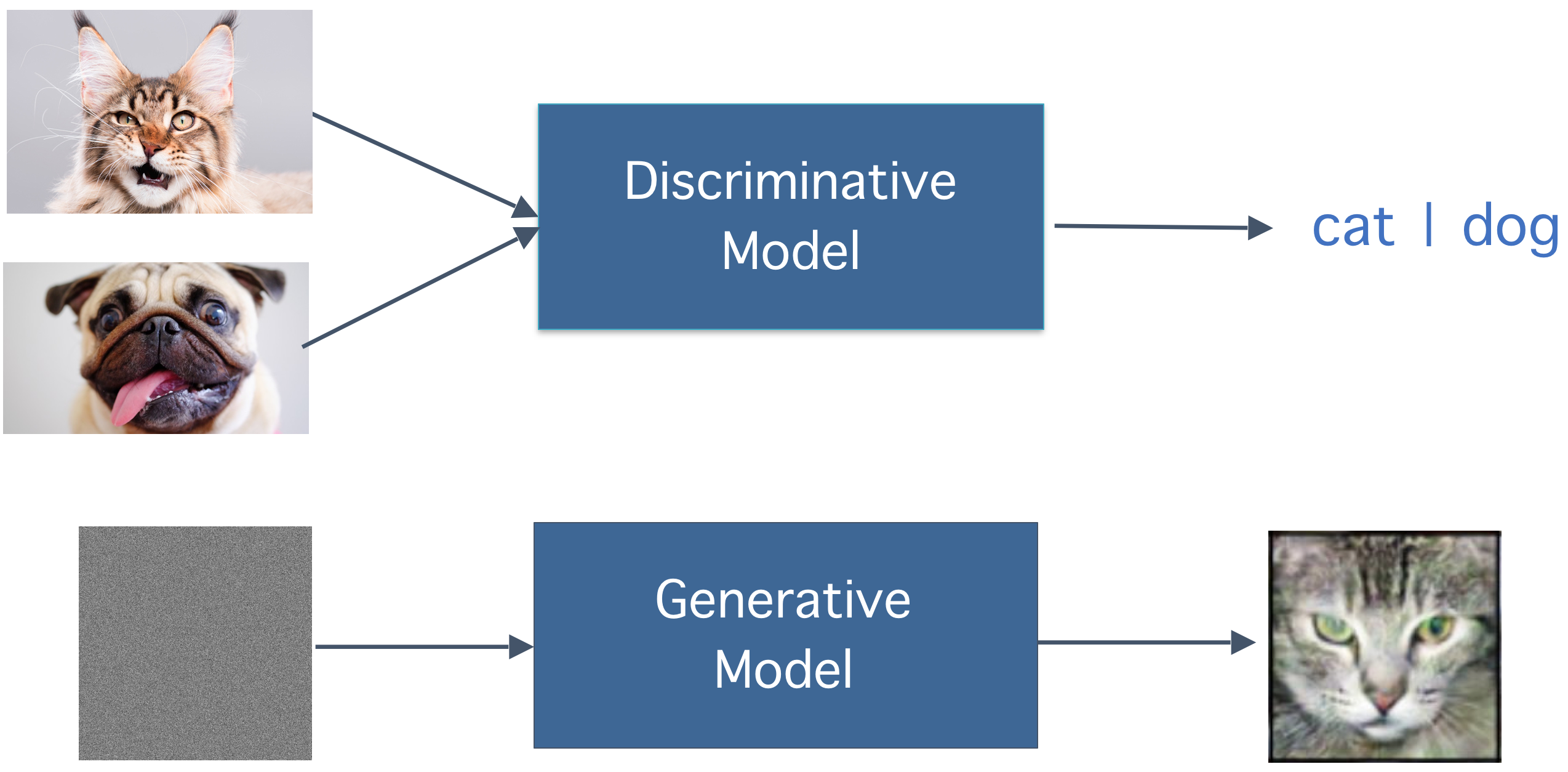}
\caption{A simple illustration of how one can use discriminative vs generative models. The former learns to distinguish between two classes, i.e., pictures of cats or dogs. The latter estimates the underlying distribution of a dataset (pictures of cats) and randomly generate realistic, yet synthetic, samples according to their estimated distribution.}
\label{fig:generative}
\end{figure}

\descr{ML approaches.} ML models can be also categorized depending on the probability distributions they learn. 
Assuming one has some input data $x$ and wants to classify it into labels $y$, then, roughly speaking, one can use either:
\begin{itemize}
\setlength\itemsep{0.1em}
\item {\em Discriminative models} to learn the conditional probability distribution $p(y|x)$, or  
\item {\em Generative models} to learn the joint probability distribution $p(x,y)$. 
Among these, Generative Adversarial Networks (GANs)~\cite{goodfellow2014generative} were originally proposed for unsupervised learning, but now also used for supervised and reinforcement learning, and have become very popular as they learn to generate new data with the same statistical properties as the training set. The two kinds of model are exemplified in Figure~\ref{fig:generative}.
\end{itemize}

\begin{figure}[t]
\centering
\includegraphics[width=0.99\columnwidth]{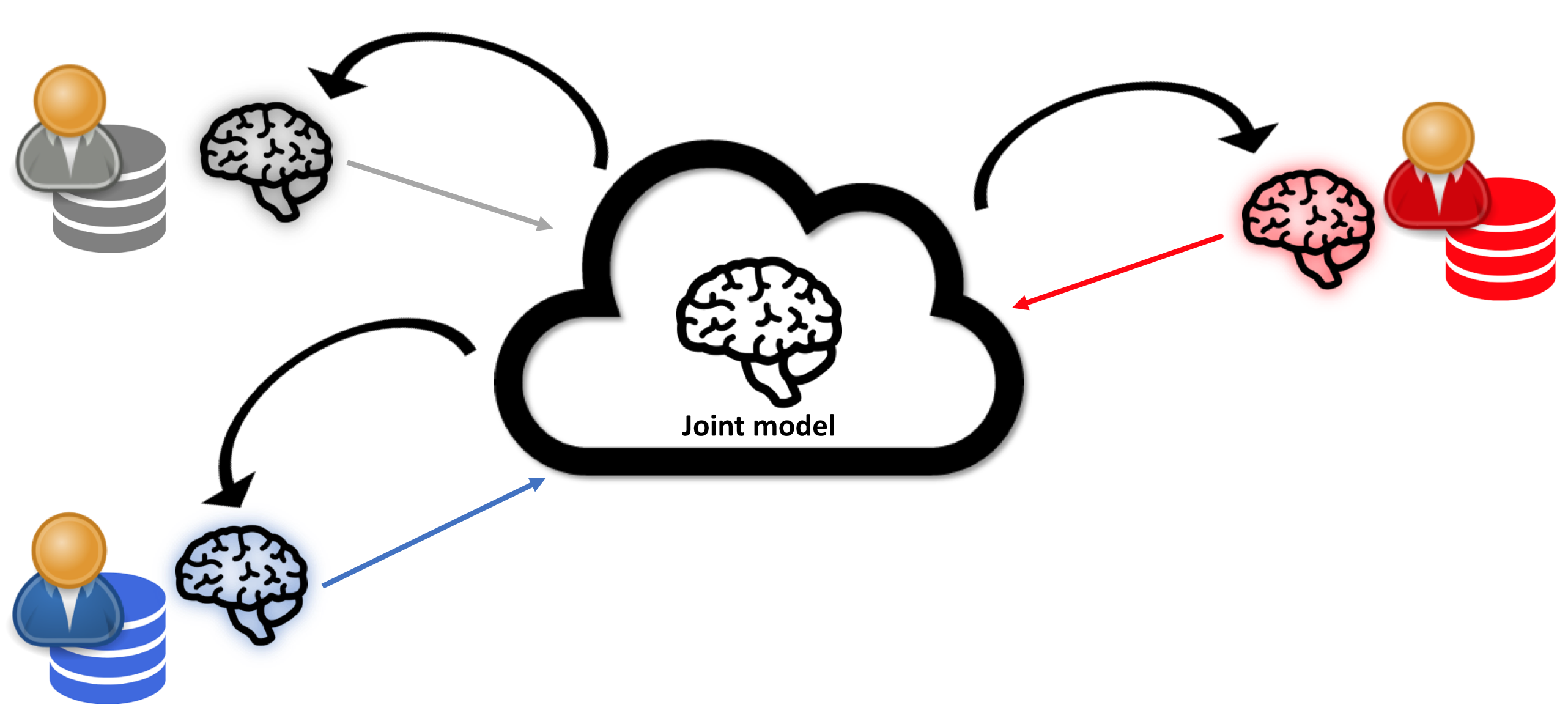}
\caption{An overview of the federated learning approach.}
\label{fig:FL0}
\end{figure}

Another distinction is based on whether the learning task is centralized or (somewhat) distributed:
\begin{itemize}
\setlength\itemsep{0.1em}
\item {\em Centralized learning:} in conventional ML methodologies, all training data is pooled and stored at a single entity, and models are trained on this joint pool. 
\item {\em Collaborative/federated learning:} multiple participants, each with their own training dataset, construct a joint model by training a local model on their own data, but periodically exchange model parameters, updates to these parameters, or partially constructed models with the other participants.
This intuition is illustrated in Figure~\ref{fig:FL0}.
There are several techniques in this category, including federated learning deployed by Google~\cite{mcmahan2016communication} on million of devices, e.g., for training of predictive keyboards on character sequences that users type on their phones.
\end{itemize}

\descr{ML Algorithms.} As discussed above, ML involves creating models. 
These can be built using different approaches, sometimes referred to as algorithms.
Supervised learning techniques include Linear Regression, Logistic Regression, K-Nearest Neighbors, Decision Trees, Support Vector Machines, Random Forest, Naive Bayes, etc.
For instance, Logistic Regression is a linear model where the probabilities describing the possible outcomes of a single trial are modeled via a logistic (logit) function; the parameters of the model are estimated with maximum likelihood estimation, using an iterative algorithm.
Unsupervised learning algorithms include clustering ones--e.g., hierarchical clustering, k-means, DBSCAN, etc.

Neural networks, which have attracted a lot of attention in recent years, are also types of ML instantiations, where learning can be either supervised or unsupervised. 
A neural net consists of a large number of simple processing nodes that are densely interconnected, usually organized into layers, to which weights are assigned. 
During training, weights are initially set to random values; as training data passes through the layers, weights are continually adjusted until training data with the same labels consistently yield similar outputs~\cite{neural}.
In particular, in deep neural networks (``deep learning''), each level learns to transform its input data into a slightly more abstract and composite representation; in fact, the word ``deep'' typically refers to the number of layers through which the data is transformed.

Overall, choosing the right machine learning algorithm depends on several factors, including, but not limited to: data size, quality and diversity, as well as what answers businesses want to derive from that data.
Additional considerations include accuracy, training time, parameters, data points and much more. 
Therefore, the right choice is typically both a combination of business need, specification, experimentation, and time available.

\subsection{Machine Learning as a Service (MLaaS)}

Many cloud providers, including Microsoft, Amazon, and IBM, have launched Machine Learning as a Service (MLaaS) offerings, aimed to help clients benefit from machine learning without the cost, time, and risk of building in-house infrastructure from scratch.
MLaaS offers ready-made, generic machine learning tools, such as predictive analytics, APIs, data visualization, and natural language processing, that can be adapted by small and medium sized companies according to their needs. 

Users who purchase MLaaS services can access these tools via prediction APIs on a pay-per-query basis. 
Typical image classification service costs around \$1--\$10 per 1,000 queries, depending on the customization and sophistication of the machine learning model.

MLaaS services vary a lot across different providers. 
In some cases, providers enable clients to download and deploy machine learning models locally, while others only allow clients to access machine learning models via a prediction query interface, which provides both the predicted label and the confidence score.
The latter is much more popular.
Finally, some platforms also allow clients to upload their own models and charge others for using their models.

\begin{figure}[t]
\centering
\includegraphics[width=0.99\columnwidth]{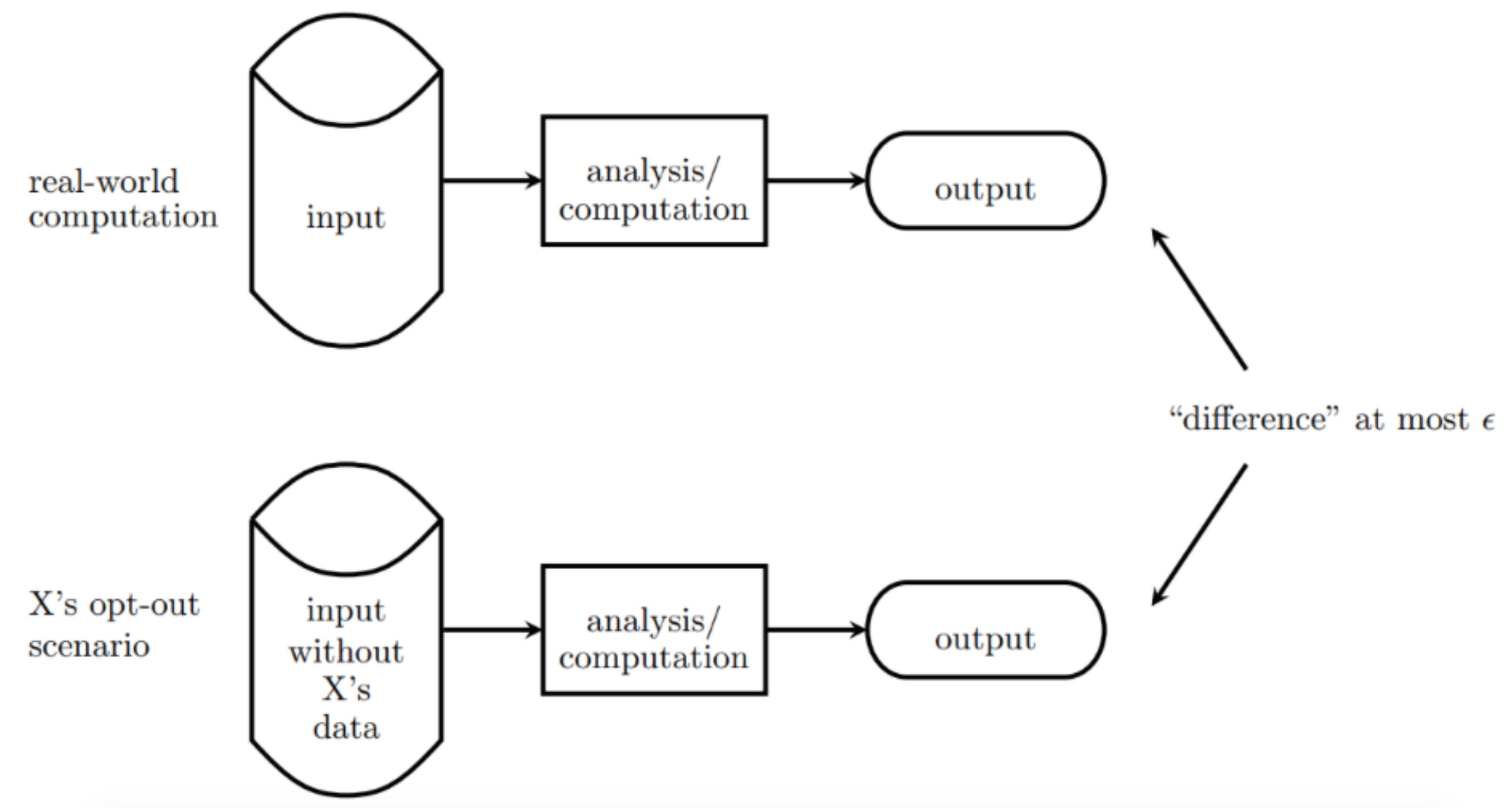}
\caption{High-level illustration of differential privacy: an adversary can distinguish between computation done on a dataset with or without data $X$ only with at most a small probability $\epsilon$.}
\label{fig:DP}
\end{figure}

\subsection{Privacy-Enhancing Technologies for Machine Learning}\label{sec:pets}

\subsubsection{Cryptography}

Cryptography, and more precisely encryption, can be used to protect data confidentiality.
There are two main primitives that are relevant in the context of ML and in general data analysis/processing:  1) {\em secure multi-party computation} (SMC), and 2) {\em fully homomorphic encryption} (FHE).

SMC allows two or more parties to jointly compute a function over their inputs, while keeping those inputs hidden from each other. 
Typically, SMC protocols build on tools like garbled circuits, secret sharing, oblivious transfer (for a detailed overview of such tools, we refer the reader to \url{https://securecomputation.org}).

Whereas, FHE is an encryption scheme that allows processing of the underlying cleartext data while it still remains in encrypted form, and without giving away the secret key.
In other words, FHE allows (almost) arbitrary computation over encrypted data.

\subsubsection{Differential Privacy (DP)}
DP addresses the paradox of learning nothing about an individual while learning useful information about a population~\cite{dwork_algorithmic_2013}. 
Generally speaking, it provides rigorous, statistical guarantees against what an adversary can infer from learning the result of some randomized algorithm. 

Typically, differentially private techniques protect the privacy of individual data subjects by adding random noise when producing statistics.
In other words, DP guarantees that an individual will be exposed to the same privacy risk whether or not her data is included in a differentially private analysis, as illustrated in Figure~\ref{fig:DP}.

\descr{Definition.} Formally, for two non-negative numbers $\epsilon, \delta$, a randomized algorithm $\mathcal{A}$ satisfies $(\epsilon, \delta)$-differential privacy if and only if, for any neighboring datasets $D$ and $D'$ (i.e., differing at most one record), and for the possible output $S \subseteq Range(\mathcal{A})$, the following formula holds: \vspace{-0.1cm}
\begin{center}
$\Pr[\mathcal{A}(D) \in S] \leq e^\epsilon \Pr[\mathcal{A}(D') \in S] + \delta$\vspace{-0.1cm}
\end{center}

\descr{The $\epsilon,\delta$ parameters.} Differential privacy analysis allows for some information leakage specific to individual data subjects, controlled by the privacy parameter $\epsilon$. This measures the effect on each individual's information on the output of the analysis. With smaller values of $\epsilon$, the dataset is considered to have stronger privacy, but less accuracy, thus reducing its utility. 
An intuitive description of $\epsilon$ privacy parameter, along with examples, is available in~\cite{nissim2017differential}.

\descr{Sensitivity.} The notion of the sensitivity of a function is very useful in the design of differentially private algorithms, and define the notion of sensitivity of a function with respect to a neighboring relationship. 
Given a query $F$ on a dataset $D$, the sensitivity is used to adjust the amount of noise required for $F(D)$. More formally, if $F$ is a function that maps a dataset (in matrix form) into a fixed-size vector of real numbers, we can define the $L_i$-sensitivity of $F$ as:
$S_i(F) = \max\limits_{D,D'} ||F(D) - F(D')||_i$,
where $||\cdot||_i$ denotes the $L_i$ norm, $i\in \{1,2\}$ and $D$ and $D'$ are any two neighboring datasets.

\section{Privacy in ML}\label{sec:privacyML}
The security of any system is measured with respect the adversarial goals and capabilities that it is designed to defend against; to this end, we now  discuss different threat models. 
Then, we attempt to provide a definition of privacy in ML, focusing on the different types of attacks, which are reviewed in detail in Section~\ref{sec:attacks}.

\subsection{Adversarial Models}

Overall, we focus on the privacy of the model.
{\em NB: adversarial examples are out of the scope of this document.}
In the rest of this section, we discuss adversarial goals related to extracting information about the model or training data. 

When the model itself represents intellectual property---e.g., in financial market systems---the model and its parameters should be kept private.
In other contexts, it is imperative that the privacy of the training data be preserved, e.g., medical applications. 
Regardless of the goal, the attacks and defenses for them relate to exposing or preventing the exposure of the model and training data.

\descr{Access.} We first discuss what kind of {\em access} the attacker might have:
\begin{itemize}
\setlength\itemsep{0.1em}
\item {\em White-Box:} she has some information about the model or its original training data, e.g., ML algorithm $h$, model parameters $\theta$, network structure, or summary, partial, or full training data.
\item {\em Black-Box:} she has no knowledge about the model. 
Rather, she might explore a model by providing a series of carefully crafted inputs and
observing outputs. 
\end{itemize}

\descr{Inference vs training.} Another variable is {\em where} the attack might take place:
\begin{itemize}
\setlength\itemsep{0.1em}
\item {\em Training Phase:} the adversary attempts to learn the model, e.g., accessing a summary, partial or all of the training data.
She might create a substitute model (aka auxiliary model) to use to mount attacks on the victim system. 
\item {\em Inference Phase:} the adversary collects evidence about the model characteristics by observing inferences made by it.

\end{itemize}

\descr{Passive vs Active.} %
Finally, one can also distinguish between passive and active attacks, roughly mirroring the traditional distinction in security literature between honest-but-curious and fully malicious adversaries.
Consider for instance federated learning, where the attacker is one of the participants in the collaborative setting: 
\begin{itemize}
\setlength\itemsep{0.1em}
\item {\em Passive attack:} the adversary passively observes the updates and performs inference, e.g., without changing anything in the training procedure;
\item {\em Active attack:} the adversary actively changes the why she operates, e.g., in the case of federated learning, by extending their local copy of the collaboratively trained model with an augmented property classifier connected to the last layer. 
\end{itemize}

\subsection{Types of attacks}\label{sec:types}
Before delving into the state of the art in actual attacks, we define what privacy means in the context of machine learning or, alternatively, what it means for a machine learning model to breach privacy.
Specific attacks proposed in literature are then reviewed in Section~\ref{sec:attacks}.

\subsubsection{Inference about members of the population}
\begin{itemize}
\setlength\itemsep{0.1em}
\item {\em Statistical disclosure:} the adversary learns something about the input to the model from the model; in theory, one would like to control statistical disclosure (this is also known as the ``Dalenius desideratum''~\cite{dalenius1977towards}), in that a model should reveal no more about the input to which it is applied than would have been known about this input without applying the model. However, as also pointed out in~\cite{shokri2017membership}, this cannot be achieved by any useful model~\cite{dwork2010difficulties}.
\item {\em Model inversion:} an adversary can use the model's output to infer the values of sensitive attributes used as input to the model. Note that it may not be possible to prevent this if the model is based on statistical facts about the population~: for example, suppose that training the model has uncovered a high correlation between a person's externally observable phenotype features and their genetic predisposition to a certain disease; this correlation is now a publicly known fact that allows anyone to infer information about the person's genome after observing that person~\cite{shokri2017membership}.
\item {\em Inferring class representatives:} overall, model inversion can be generalized to potential breaches where the adversary, given some access to the model, infers features that characterize each class, making it possible to construct representatives of these classes. 

\end{itemize}

\subsubsection{Inference about members of the training dataset}
Here the focus is on privacy of the individuals whose data was used to train the model. This motivation is closely related to the original goals of differential privacy~\cite{dwork2008differential}.
Of course, members of the training dataset are members of the population, too. 
Therefore, one should focus on what the model reveals about them beyond what it reveals about an arbitrary member of the population: 
\begin{itemize}
\setlength\itemsep{0.1em}
\item {\em Membership inference:} given a model and an exact data point, the adversary infers whether this point was used to train the model or not.
\item {\em Property inference:} training data may not be identically distributed across different users whose records are in the training set; unlike model inversion, the adversary tries to infer properties that are true of a {\em subset} of the training inputs but not of the class as a whole.  For instance, when Bob's photos are used to train a gender classifier, she infers that Alice appears in some of the photos. 
\end{itemize}

\subsubsection{Inferring Model Parameters}\label{sec:stealing}

As discussed earlier, MLaaS allows model owners to charge others for queries to their commercially valuable models. 
This pay-per-query deployment option exemplifies an increasingly common tension: on the one hand, the query interface of an ML model may be widely accessible, yet the model itself and the data on which it was trained may be proprietary and confidential. 
Moreover, for security applications such as spam or fraud detection, an ML model's secrecy is critical to its utility; an adversary that can learn the model can also often evade detection~\cite{ateniese2015hacking}.

In this space, we can distinguish between:
\begin{itemize}
\setlength\itemsep{0.1em}
\item {\em Model Extraction:} a black-box adversary that can query an ML model to obtain predictions on input feature vectors, and may or may not know the model type (e.g., logistic regression) or the distribution over the data used to train the model. 
The adversary's goal is to extract an equivalent or near-equivalent ML model.
\item {\em Functionality Stealing:} Rather than stealing the model, here the ultimate goal is to create ``knock-offs'' of the (black-box) model solely based on input-output pairs observed from MLaaS queries~\cite{orekondy2019knockoff}.

\end{itemize}

\section{Attacks}\label{sec:attacks}

\subsection{Membership Inference Attack (MIA)}\label{sec:inf_att}
As mentioned above, membership inference relates to the problem of deciding, given a data point, whether or not it was included in the training dataset. 

\subsubsection{Definition and Relevance}
Formally, in a membership inference attack (MIA), the adversary, given a target datapoint $d^*$ and {\em some} access to a model $h_\theta(x)$, tries to infer whether $d^*\in x$.
This can constitute a serious privacy breach in a number of settings, which we discuss next.

\descr{Sensitivity of task/model.} First of all, MIA can directly violate privacy if inclusion in a training set is itself sensitive based on the nature of the task at hand. 
For example, if health-related records (or images like MRIs) are used to train a classifier, discovering that a specific record was used for training inherently leaks information about the individual's health.  
Similarly, if images from a database of criminals are used to train a model predicting the probability that one will re-offend, successful membership inference exposes an individual's criminal history. 

\descr{Signal of leakage.} When a record is fully known to the adversary, learning that it was used to train a particular model is an indication of information leakage through the model.
Overall, MIA is often considered to be a signal---a measuring stick of sort---that access to a model leads to potentially serious privacy breaches.
In fact, MIAs are often a gateway to further attacks: e.g., if the adversary infers that data of a victim is part of the information she has access to she can mount other attacks, like profiling, property inference, etc.

\descr{Establishing wrongdoing.} On the other hand, MIA can also be used by regulators to support the suspicion that a model was trained on personal data without an adequate legal basis, or for a purpose not compatible with the data collection.
For instance, DeepMind was recently found to have used personal medical records provided by the UK's National Health Service for purposes beyond direct patient care; the basis on which the data was collected~\cite{verge}.

\descr{MIA beyond machine learning.} As a side note, we remark that MIAs have been studied not only in the context of machine learning, but also in other fields.
Overall, given a data point $d^*$ and a function $f(x)$, one can define membership inference as the problem of determining whether $d^*$ is part of the input to the function $f$, i.e., $d^* \in x$.
Often, $f$ is some aggregation function, and in fact researchers have demonstrated the existence of successful MIAs against aggregate statistics in the context of genomic studies~\cite{homer2008resolving}, location data~\cite{pyrgelis2017knock}, and noisy statistics in general~\cite{dwork2015robust}.

\subsubsection{State of the Art}

\descr{Attacking Machine Learning as a Service.} MIA against black-box machine learning models was first studied 
by Shokri et al.~\cite{shokri2017membership}, in the context of supervised learning.
They focus on classification models trained by commercial Machine Learning as a Service (MLaaS) providers, such as Google and Amazon, whereby a user has API access to a trained model.

\begin{figure}[t]
\centering
\includegraphics[width=0.99\columnwidth]{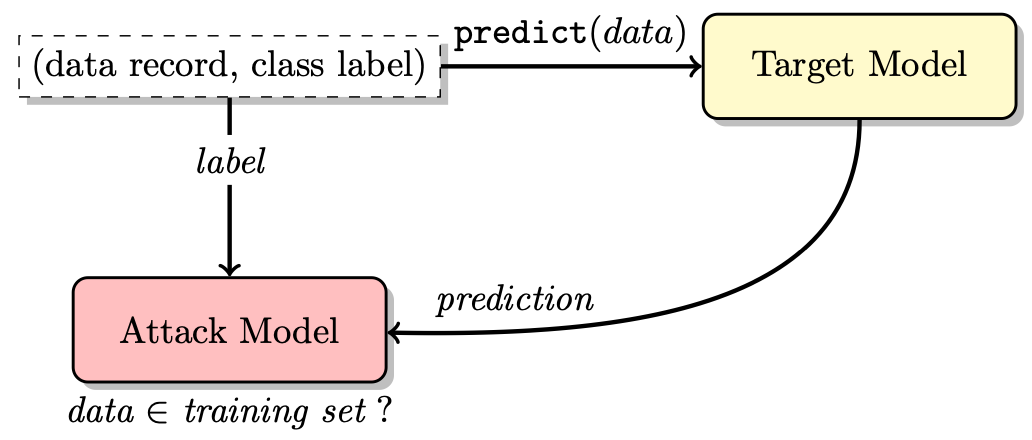}
\caption{Membership inference attack against MLaaS in the black-box setting~\cite{shokri2017membership}.}
\label{fig:MIA1}
\end{figure}

More specifically, customers in possession of a dataset and a data classification task can upload the dataset to the MLaaS service and pay it to construct a model. 
The service then makes the model available to the customer---typically as a black-box API. 
This setting is illustrated in Figure~\ref{fig:MIA1}.
For example, a mobile-app maker can use such a service to analyze users' activities and query the resulting model inside the app to promote in-app purchases to users when they are most likely to respond. 
Moreover, some machine-learning services also let data owners expose their models to external users for querying or even sell them.

\descr{Inference via overfitting.} Shokri et al.~\cite{shokri2017membership}'s approach exploits differences in the model's response to inputs that were or were not seen during training. 
For each class of the targeted black-box model, they train a \emph{shadow model}, with the same machine learning technique; the intuition is that the model ends up ``overfitting'' on data used for training~\cite{shokri2017membership}.
Overfitting is a modeling error that occurs when a function is too closely fit to a limited set of data points, and thus performs better on the training inputs than on the inputs drawn from the same population but not used during the training.
Therefore, the attacker can exploit the confidence values on inputs belonging to the same classes and learn to infer membership.

Then, Salem et al.~\cite{salem2018ml}, although also building on overfitting signals, relax several adversarial assumptions---e.g., using multiple shadow models, knowledge of the target model structure, and having a dataset from the same distribution as the target model's training data---showing that MIAs are very broadly applicable at low cost and thereby pose a more severe risk than previously thought.
Overall, the attacks are relatively successful across the board, more or less so depending on the adversary's prior knowledge, the datasets, the task at hand, etc.

\descr{ML and overfitting.} Another line of work studies the relationship between overfitting and information leakage.
In particular, Yeom et al.~\cite{yeom2017unintended} assume that the adversary knows the distribution from which the training set was drawn and its size, and that the adversary colludes with the training algorithm.
Their attacks are close in performance to Shokri et al.'s~\cite{shokri2017membership}, and show that, besides overfitting, the influence of target attributes on model's outputs also correlates with successful attacks.
Truex et al.~\cite{demyst2018} extend~\cite{shokri2017membership} to a more general setting and show how membership inference attacks are largely transferable. 

\begin{figure*}[t]
\centering
\includegraphics[width=1.4\columnwidth]{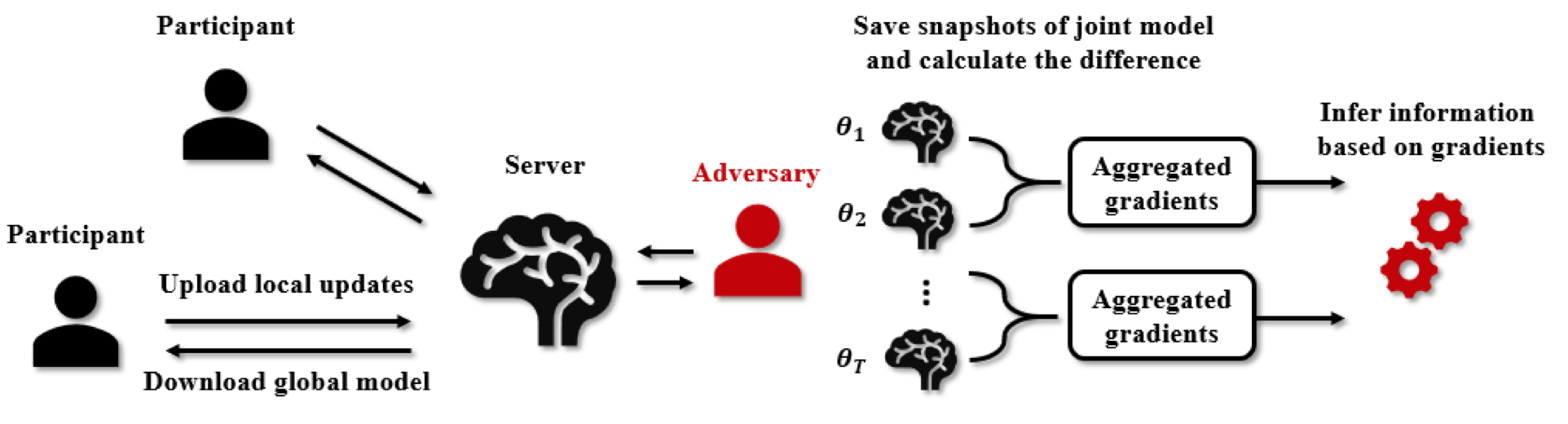}
\vspace{-0.2cm}
\caption{Inference attacks against federated learning (passive adversary) by Melis et al.~\cite{melis2019exploiting}.}
\label{fig:FL1}
\end{figure*}

\descr{Generative models.} While the research discussed above focus on discriminative models, other work target generative models.
As discussed in Section~\ref{sec:ML}, they are used to generate new samples from the same underlying distribution of a given training dataset, e.g., to artificially generate plausible images and videos.
Here the attacker targets a MLaaS engine that provides synthetic samples on demand -- e.g., the user's query is {\em ``provide an image sample of a cat''} -- based on a trained generative model. 
Once again, inferring whether specific data points are part of the training set for that generative model may constitute a serious privacy breach.
Note that membership inference on generative models is much more challenging than on discriminative models: in the former, the attacker cannot exploit confidence values on inputs belonging to the same classes, and therefore it is more difficult to detect overfitting and mount the attack.

Hayes et al.~\cite{hayes2019logan} consider both black-box and white-box attacks: in the former, the adversary can only make queries to the model under attack, i.e., the target model, and has no access to the internal parameters. 
In the latter, she also has access to the parameters. 
To mount the attacks, they train a Generative Adversarial Network (GAN)~\cite{goodfellow2014generative} on samples generated from the target model; i.e,, using generative models as a method to learn information about the target generative model, and thus create a local copy of the target model from which they can launch the attack. 
The intuition is that, if a generative model overfits, then a GAN---which combines a discriminative model and a generative model---should be able to detect this overfitting, since the discriminator is trained to learn statistical differences in distributions. 
Moreover, for white-box attacks, the attacker-trained discriminator itself can be used to measure information leakage of the target model.

\descr{Federated Learning.} In this setting, the attack can be mounted by an adversary, a participant in the federated learning, attempting to infer whether a specific record is part of the training set of either a specific or any participant.
The first MIA against federated learning is presented by Melis et al.~\cite{melis2019exploiting}, whose main intuition is to exploit unintended leakage from either the {\em embedding layer} (all deep learning models operating on non-numeric data where the input space is discrete and sparse first use an embedding layer to transform inputs into a lower-dimensional vector representation) or the {\em gradients} (in deep learning models, gradients are computed by back-propagating the loss through the
entire network from the last to the first layer).
An illustration of Melis et al.~\cite{melis2019exploiting}'s attack is in Figure~\ref{fig:FL1}.
Then, Nasr et al.~\cite{nasr2019comprehensive} design MIAs during the training phase in a white-box setting, including passive and active attackers based on the different adversary prior knowledge.

\subsection{Model Inversion}
As mentioned earlier, model inversion techniques aim to infer class features and/or construct class representatives, given that the adversary has {\em some} access (either black-box or white-box) to a model.

\subsubsection{Definition and Early Work} 
The concept of model inversion is introduced by Fredrikson et al.~\cite{fredrikson2014privacy,fredrikson2015model}.
First, they show how an attacker can rely on outputs from a classifier to infer sensitive features used as inputs to the model itself: given the model and some demographic information about a patient whose records are used for training, an attacker might predict sensitive attributes of the patient.
Then, they use so-called ``hill-climbing'' on the output probabilities of a computer-vision classifier to reveal individual faces from the training data.

These techniques are sometimes described as violating privacy of the training data, even though the inferred features characterize an entire class and not specifically the training data, except in the cases of pathological overfitting where the training sample constitutes the entire membership of the class.

\subsubsection{Further Attacks}
\descr{Collaborative learning.} Hitaj et al.~\cite{hitaj2017deep} show that a participant in collaborative learning can use GANs to construct class representatives.
However, this technique has been evaluated only on models where all members of the same class are visually similar (handwritten digits and faces). 
Thus, there is no evidence that it produces actual training images or can distinguish a training image and another image from the same class.

Aono et al.~\cite{aono2017privacy} show that, in the collaborative deep learning protocol of~\cite{shokri2015privacy}, an adversarial server can partially recover participants' data points from the shared gradient updates, although in a greatly simplified setting where the batch consists of a single data point.  

\descr{Unintended Memorization.} Song et al.~\cite{song2017machine} engineer a machine learning model that memorizes the training data, which can then be extracted with black-box access to the model, without affecting the accuracy of the model on its primary task.

Then, Carlini et al.~\cite{carlini2018secret} show that deep learning-based generative sequence models trained on text data can unintentionally
memorize specific training inputs, which can then be extracted with black-box access.
Even though the models are trained on text, extraction
is demonstrated only for sequences of digits (artificially introduced into the text), which are not affected by the relative word frequencies
in the language model.

\begin{figure}[t]
\centering
\includegraphics[width=0.15\textwidth]{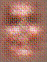}~
\includegraphics[width=0.15\textwidth]{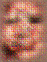}\\[1ex]
\includegraphics[width=0.15\textwidth]{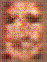}~
\includegraphics[width=0.15\textwidth]{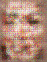}
\caption{Samples from a GAN attack on a gender classification model where the class is ``female.''}\label{ganimage}
\end{figure}

\subsection{Property Inference}
As mentioned above, work presented in~\cite{fredrikson2015model, hitaj2017deep, ateniese2015hacking} aimed to infer properties that characterize an entire class: for example, given a face recognition model where one of the classes is Bob, infer what Bob looks like (e.g., Bob wears glasses).
However, while Ateniese et al.~\cite{ateniese2015hacking} are actually the first, to the best of our knowledge, to reason about extracting ``something meaningful relating to properties of the training set,'' it is not clear that hiding this kind of information in a good classifier is possible or desirable.

\subsubsection{Attacks}

\descr{Melis et al.~\cite{melis2019exploiting}'s work.} By contrast, here we focus on the adversarial goal of inferring properties that are \emph{true of a subset of the training inputs but not of the class as a whole}.  
For instance, when Bob's photos are used to train a gender classifier, can the attacker infer that Alice appears in some of the photos?
In particular, Melis et al.~\cite{melis2019exploiting} focus on the properties that are \emph{independent} of the class's characteristic features.
In contrast to the face recognition example, where ``Bob wears glasses'' is a characteristic feature of an entire class, in their gender classifier study they infer whether people in Bob's photos wear glasses\textemdash even though wearing glasses has no correlation with gender.  
There is no ``legitimate'' reason for a model to leak this information; it is purely an artifact of the learning process.

The work in~\cite{melis2019exploiting} studies this kind of property inference in the context of collaborative/federated learning. 
More specifically, their intuition is that a participant's contribution to each iteration of collaborative learning is based on a batch of their training data, and the adversary can infer \emph{single-batch properties}, i.e., detect that the data in a given batch has the property but other batches do not.  
She can also infer \emph{when a property appears in the training data}, which has very serious privacy implications.  
For instance, the adversary can infer when a certain person starts appearing in a participant's photos or when the participant starts visiting a certain type of doctors.
Finally, they infer properties that characterize a participant's entire dataset (but not the entire class), e.g., authorship of the texts used to train a sentiment-analysis model.

\descr{Follow-up work.} Ganju et al.~\cite{ganju2018property} also develop property inference attacks, mainly against fully connected, relatively shallow neural networks. They focus on the post-training, white-box release of models trained on sensitive data, and the properties inferred by the adversary may or may not be correlated with the main task.

\subsection{Model and Functionality Stealing}
Finally, we look into adversarial efforts toward inferring model parameters.

\subsubsection{Model Extraction}
The concept of model stealing, or extraction, is first presented by Tramer et al.~\cite{tramer2016stealing}.
In this kind of attack, an adversary with black-box access, but no prior knowledge of an ML model's parameters or training data, aims steal the model parameters.
The intuition behind their attack is to exploit the information-rich outputs returned by the ML prediction APIs, e.g., high-precision confidence values in addition to class labels. %

Consider the case of ML algorithms like logistic regression: the confidence value is a simple log-linear function $1/(1+e^{-(w\cdot x+\beta)})$ of the $d$-dimensional input vector $x$. By querying $d + 1$ random $d$-dimensional inputs, an attacker can with high probability solve for the unknown $d + 1$ parameters $w$ and $\beta$ defining the model.
(Such equation-solving attacks extend to multi-class logistic regressions and neural networks).

Overall, Tramer et al.~\cite{tramer2016stealing}'s work is focused on inferring model parameters. 
In follow-up work, other researchers have then gone a step further and perform hyperparameter stealing~\cite{wang2018stealing}, architecture extraction~\cite{joon2018towards}, etc.
In the former, the focus is on hyperparameters, rather than parameters, which are configurations external to the model and whose values cannot be estimated from data.
In the latter, a black-box adversary succeeds to infer (hidden) model architectures (e.g., the type of non-linear activation) of neural networks in MLaaS as well as their optimization processes (e.g., stochastic gradient descent or ADAM).

\subsubsection{Functionality Extraction} 
As mentioned in Section~\ref{sec:types}, the goal of functionality extraction is, rather than to steal the model, to create ``knock-offs.'' 
In~\cite{orekondy2019knockoff}, Orekondy et al.~do so solely based on input-output pairs observed from MLaaS queries. %
More specifically, the adversary interacts with a black-box ``victim'' Convolutional Neural Network (CNN) by providing it input images and obtaining respective predictions. 
The resulting image-prediction pairs are used to train a knock-off model, e.g., to compete with the victim model at the victim's task. 

Additional work in this category includes~\cite{papernot2016practical}, whereby the adversary trains a local model to substitute for a victim deep neural network (DNN), using inputs synthetically generated by an adversary and labeled by the target DNN.

\subsection{Take Aways}

Throughout this section, we have presented and discussed the state of the art in attacks against privacy in ML.

\descr{Membership inference attacks are real.} As evident from the above discussion, there has been a very significant amount of research work on membership inference attacks against ML.
Arguably, this is motivated by 1) the seriousness of the privacy risks stemming from such attacks, 2) the fact that MIA is often just a signal of leakage and can serve as a canary for broad privacy issues, and 3) the interesting challenges in making attacks more effective, less reliant on strong assumptions, etc.

Overall, several attacks have been proposed in the context of a wide variety of datasets (images, text, etc.), models (discriminative, generative, federated), as well as threat models (API access, white-box, black-box, active, passive, etc.). 
Such attacks are realistic but obviously their effectiveness depends on the actual settings, e.g., adversary's knowledge of records, model parameters, etc., and are likely to affect certain users more than others.

While we discuss possible defenses and the overall outlook, respectively, in Section~\ref{sec:defenses} and~\ref{sec:discussion}, we are confident in arguing that MIAs are a real problem that at the very least should make practitioners and researchers question whether deploying ML models in the wild is a good idea, privacy-wise, whenever training data is sensitive.
However, further work is needed to provide clear guidelines and usable tools for practitioners willing to provide access to trained model to fully understand the privacy risks, on their specific data/specific learning task, for the users whose data is used for training.

\descr{Limitations of model inversion.} Although research roughly falling in the ``model inversion'' category is important, %
we believe there are some limitations in what they mean for privacy. %
Class members produced by model inversion and GANs are similar to the training inputs only if all members of the class are similar, as is the case for MNIST (the dataset of handwritten digit used in~\cite{hitaj2017deep}) and facial recognition.
This does not violate privacy of the training data; it simply shows that machine learning works as it should. 
A trained classifier reveals the input features characteristic of each class, thus enabling the adversary to sample from the class population. 
For instance, Figure~\ref{ganimage} shows GAN-constructed images for the gender classification task on the Labeled Faces in the Wild (LFW) dataset, taken from~\cite{melis2019exploiting}.
These images show a generic female face, but there is no way to tell from them whether an image of a specific female was used in training or not.

Therefore, the informal property violated by such attacks %
is, roughly speaking: ``a classifier should prevent users from generating an input that belongs to a particular class or even learning what such an input looks like.''
However, it is not clear why this property is desirable or whether it is even achievable.
In fact, this motivated us to study what we defined as property inference attacks. %

\descr{Property inference needs further work.}
Overall, property inference attacks are not to be ignored, even though, their effectiveness depends on the context.
As mentioned earlier, inferring sensitive attributes is really a privacy breach when the attacker can confidently assess that those attributes related to records in the training set, and even more so if they do not leak simply because the class the model is learning to classify is strictly correlated.

So really the only ``attack'' in this sense we are aware of is that of Melis et al.~\cite{melis2019exploiting}, which has only been studied in the context of collaborative learning.
Even in that case, the authors essentially show that the accuracy of the attack quickly degrades with increasing number of participants.
In fact, if this is large enough, then differentially private defenses based on the moments accountant method~\cite{abadi2016deep} (discussed in Section~\ref{sec:defenses}) can be used to thwart such attacks.

It remains however an open research questions to investigate whether property inference attacks: 1) are possible, as per our definition, in non collaborative learning settings and at scale, and 2) can be thwarted in collaborative settings involving a small number of participants. 
(We discuss the latter in more detail Section~\ref{sec:defenses}.)

\section{Defenses}\label{sec:defenses} 

Overall, we can categorize defenses against attacks discussed in this document and in general aiming to protect confidentiality and privacy in ML based on the main tools they rely on.
These include advanced {\em privacy-enhancing technologies} like cryptography and differential privacy (reviewed in Section~\ref{sec:pets}) as well approaches used as part of the learning process (mainly, training) to reduce information available to the adversary.

\subsection{Cryptography} 
Cryptography in ML can support confidential computing scenarios where, for instance, a server has a model trained on its private data and wishes to provide inferences (e.g., classification) on clients' private data. 
In this context, there are a number of research proposals and prototypes in literature, which allow the client to obtain the inference result without revealing their input to the server, while at the same preserving the confidentiality of the server's model.
For instance, privacy-enhancing tools based on secure multi-party computation (SMC) and fully homomorphic encryption (FHE) have been proposed to securely train supervised machine learning models, such as matrix factorization~\cite{nikolaenko2013privacy}, linear classifiers~\cite{bos2014private,graepel2013ml}, decision trees~\cite{bost2014machine,lindell2000privacy}, linear regressors~\cite{du2004privacy}, and neural networks~\cite{bonawitz2017practical,dowlin2016cryptonets,liu2017oblivious,mohassel2017secureml}.

SMC has also been used to build privacy-preserving neural networks in a distributed fashion.
For instance, SecureML~\cite{mohassel2017secureml} starts with the data owners (clients) distributing their private training inputs among two non-colluding servers during the setup phase; the two servers then use MPC to train a global model on the clients' encrypted joint data.
Then, Bonawitz et al.~\cite{bonawitz2017practical} use secure multi-party aggregation
techniques, tailored for federated learning, to let participants encrypt their updates so that the central parameter server only recovers the sum of the updates.

\descr{Confidentiality vs Privacy.} Overall, cryptography in ML is really aimed at protecting {\em confidentiality}, rather than {\em privacy}, which constitutes the main focus of our report.
The two terms are often confused, both in the context of ML and in general, but they actually refer to different properties.
Confidentiality is an explicit design property whereby one party wants to keep information (e.g., training data, testing data, model parameters, etc.) hidden from both the public and other parties (e.g., clients with respect to servers or vice-versa).
Whereas, privacy is about protecting against {\em unintended} information leakage, whereby an adversary aims to infer sensitive information through some (intended) interaction with the victim. 
In other words, cryptographically-enforced confidential computing does not provide any guarantees about what the output of the computation reveals.
Therefore, we will focus on privacy rather than confidentiality defenses.

\begin{figure*}[t]
\centering
\includegraphics[width=1.6\columnwidth]{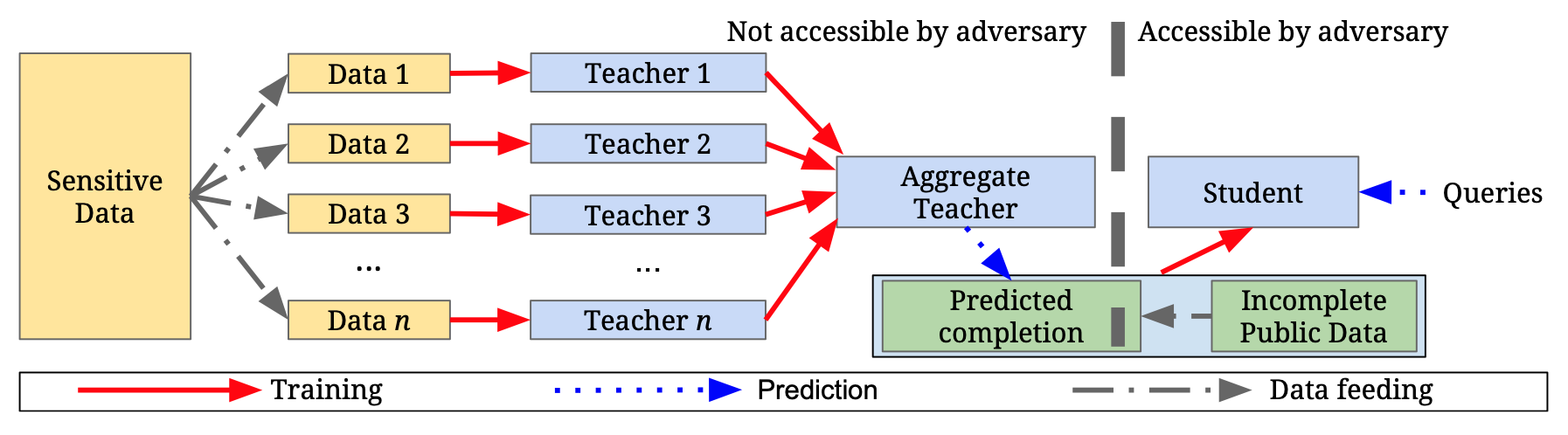}
\vspace{-0.4cm}
\caption{Overview of PATE's approach~\cite{papernot2018scalable}.}
\label{fig:pate}
\end{figure*}

\subsection{Differential Privacy (DP)} 
The state-of-the-art method for providing access to information free from inferences is to satisfy differential privacy (DP)~\cite{dwork2008differential}.
This applies to ML as well, and more precisely to providing access to models that have been trained on (sensitive) datasets~\cite{abadi2016deep,phan2017preserving,shokri2015privacy,papernot2016semi,papernot2018scalable}.
More precisely, there are two main privacy-preserving model-training approaches  in literature: 1) using noisy stochastic gradient descent (noisy SGD)~\cite{abadi2017protection}, and 2) Private Aggregation of Teacher Ensembles (PATE)~\cite{papernot2016semi,papernot2018scalable}.

\descr{Noisy SGD.} As mentioned in Section~\ref{sec:ML}, ML models are often used trained using stochastic gradient descent (SGD). The intuition is to add noise to the SGD process; however, the main challenge is to do so while ensuring that the noise is carefully calibrated. 
The sensitivity of the final value of $\theta$ (the parameter vector) to the elements of the training data is generally hard to analyze. 
On the other hand, since the training data affects $\theta$ only via the gradient computations, one may achieve privacy by bounding gradients (by clipping) and by adding noise to those computations. 
In a nutshell, this is the idea behind the seminal work by Abadi et al.~\cite{abadi2016deep}, in particular in the accounting of the privacy loss, and hence it is general referred to as the {\em Moments Accountant} algorithm.

\descr{PATE.} To protect the privacy of training data during learning, PATE transfers knowledge from an ensemble of ``teacher'' models trained on partitions of the data to a ``student'' model; see Figure~\ref{fig:pate}.
Intuitively, privacy is provided by training teachers on disjoint data, and strong guarantees stem from noisy aggregation of teachers' answers.

\descr{Collaborative Learning.}
In the collaborative learning setting, Shokri and Shmatikov~\cite{shokri2015privacy} support distributed training of deep learning networks in a privacy-preserving way. 
Specifically, their system relies on the input of independent entities which aim to collaboratively build a machine learning model without sharing their training data. 
To this end, they selectively share subsets of noisy model parameters during training. 
Moreover, federated learning proposals, which have already mentioned, tackle the problem of training deep learning models with differential privacy guarantees for the tasks of training language models~\cite{mcmahan2017learning} and digits classification~\cite{geyer2017differentially}.

\subsection{Trusted Execution Environments} %
A different line of work focuses on privacy (as well as integrity) guarantees for ML computations in untrusted environments (i.e., tasks outsourced by a client to a remote server, including MLaaS) by leveraging so-called Trusted Execution Environments (TEEs), such as Intel SGX or ARM TrustZone.
TEEs use hardware and software protections to isolate sensitive code from other applications, while attesting to its correct execution. 
The main idea is that TEEs outperform purely cryptographic approaches by multiple orders of magnitude.

In this area, there are three main approaches. 
The first includes work supporting oblivious data access patterns~\cite{ohrimenko2016oblivious} and in general training for a range of ML algorithms run inside SGX~\cite{hunt2018chiron,hynes2018efficient}.
The second, by Tramer and Boneh~\cite{tramer2018slalom}, focuses on {\em high performance} execution of Deep Neural Networks (DNNs) in TEEs, by efficiently partitioning DNN computations between trusted and untrusted devices.
The third, by Hanzlik et al.~\cite{hanzlik2018mlcapsule}, is essentially a guarded offline deployment of MLaaS: models are executed locally on the client's side (therefore, the data never leaves the device).

\subsection{ML-Specific Approaches}
Finally, a number of ML techniques are used to reduce information available to the adversary to mount their attacks.
For instance, {\em dropout} is a popular technique often used to mitigate overfitting in neural networks; as such, this might reduce the effectiveness of MIAs based on overfitting.

Additional techniques in this space include weight normalization (re-parameterization of the weights vectors that decouples the length of those weights from their direction), dimensionality reduction (e.g., only using inputs that occur many times in the training data), selective gradient sharing (in collaborative learning, participants could share only a fraction of their gradients during each update), etc.

\section{Conclusion}
This document provided a literature review on privacy and machine learning.
We provided ample background information on relevant concepts, including machine learning, differential privacy, etc., as well as adversarial models.
Then, we presented a wide range of attacks that relate to private and/or sensitive information leakage.
Finally, we discussed recent results attempting to defend against such attacks. 
Next, we list three areas where further work is desperately needed.

\subsection{Severity of Attacks} 
Overall, our review showed that there exists a large body of literature presenting a wide range of different attacks. 
Specifically, we covered membership inference attacks (MIAs), model inversion, property inference, as well as model and functionality stealing attacks.

We concluded that membership inference attacks are very much possible, but it is hard to grasp the real-world effect on actually deployed models due to the lack of case studies vis-a-vis impact on actual users, relation to adversary's prior knowledge, etc.
We also offered a critic of model inversion attacks, while covering less studied attacks in the field of property inference and model stealing.
Much work is left to be done -- especially considering ways to provide guidelines and evaluation framework for practitioners.

\subsection{Policy Implications and Further Study Needed} 
The implication of the attacks covered in this manuscript vis-a-vis policy and data protection is also largely unexplored.
The only exception in this context is the work by Cohen and Nissim~\cite{cohen2019towards}, which rephrases privacy attacks in the General Data Protection Regulation (GDPR) framework and more specifically within the {\em ``singling out''} concept.
While the GDPR heavily focuses on the concept of identification, what it means for a person to be ``identified, directly or indirectly'' is not clear.
As pointed in~\cite{cohen2019towards}, Recital 26 sheds a little more light: ``To determine whether a natural person is identifiable account should be taken of all the means reasonably likely to be used, such as singling out, either by the controller or by another
person to identify the natural person directly or indirectly.'' 

Therefore, singling out is one way to identify a person in data, and only data that does not allow singling out may be excepted from the regulation.
Clearly, more work linking up privacy attacks (and defenses) with regulation and data protection efforts needs to ramp up.

\subsection{Need for Better Evaluations} 
Overall, several defense techniques against privacy attacks have been proposed over the past few years.
However, it is very hard to assess how generalizable they are and what is the trade-off they incur with respect to privacy and utility.
This prompts the need for a more thorough evaluation of how defenses fare in practice, vis-a-vis realistic use cases and datasets, rather than the standard public ones that, more often than not, say little or nothing about real-world performance.

In this context, some recent work has taken some good steps in the right direction; for instance,  Jayaraman and Evans~\cite{jayaraman2019evaluating} study the impact of variable choices of the $\epsilon$ parameter, different variants of differential privacy, and several learning tasks on both utility and privacy (including in the context of MIAs) for privacy-preserving machine learning. 
Alas, however, their main finding is that there is no way to obtain privacy for free?relaxed definitions of differential privacy that reduce the amount of noise needed to improve utility also increase the privacy leakage. 
In other words, current mechanisms for differentially private machine learning rarely offer acceptable utility-privacy trade-offs for complex learning tasks: settings that provide limited accuracy loss provide little effective privacy, and settings that provide strong privacy result in useless models.

Once again, this points to the need to better understand where trade-offs are possible, in what context, and at what expenses, rather than hoping to deploy generic, one-size-fits-all defenses across the board.

\descr{Ackwnoledgments.} The author wishes to thanks Vitaly Shmatikov, Luca Melis, Jamie Hayes, and Yang Zhang for valuable discussions and feedback.

{\small
\bibliographystyle{abbrv}
%\bibliography{bibliography}

}
\end{document}